# UTSig: A Persian Offline Signature Dataset

Amir Soleimani [1,*], Kazim Fouladi [1], Babak N. Araabi [1,2,*]

[1] Machine Learning and Computational Modeling Lab, Control and Intelligent Processing Centre of Excellence, School of Electrical and Computer Engineering, University of Tehran, Tehran, Iran
[2] School of Cognitive Sciences, Institute for Research in Fundamental Sciences (IPM), Tehran, Iran
[*] a.soleimani.b@gmail.com, araabi@ut.ac.ir

**Abstract:**
The pivotal role of datasets in signature verification systems motivates researchers to collect signature samples. Distinct characteristics of Persian signature demands for richer and culture-dependent offline signature datasets. This paper introduces a new and public Persian offline signature dataset, UTSig, that consists of 8280 images from 115 classes. Each class has 27 genuine signatures, 3 opposite-hand signatures, and 42 skilled forgeries made by 6 forgers. Compared with the other public datasets, UTSig has more samples, more classes, and more forgers. We considered various variables including signing period, writing instrument, signature box size, and number of observable samples for forgers in the data collection procedure. By careful examination of main characteristics of offline signature datasets, we observe that Persian signatures have fewer numbers of branch points and end points. We propose and evaluate four different training and test setups for UTSig. Results of our experiments show that training genuine samples along with opposite-hand samples and random forgeries can improve the performance in terms of equal error rate and minimum cost of log likelihood ratio.

## 1. Introduction

Signature is one of the most widespread personal attributes for authentication. It is simple, cheap and acceptable to people, official organizations and courts. However, Signature Verification Systems (SVSs) suffer from variables that affect the performance, such as writing instrument, paper, and physical condition of the writer. On the other head, an accurate SVS demands for considerably large number of samples.

SVSs aim to help Forensic Handwriting Experts (FHEs) in decision making. In the literature, SVSs based on acquisition approach divide into offline and online categories. In offline systems, signatures are 2-D images, while in online systems samples are described by position, velocity, pen orientation, and pressure sequences [1–5]. The abundance and uniqueness of information in online systems provide more accurate results [2]; however, online mode is not completely natural for the users [6].

Signatures based on authenticity divide into three categories: genuine (authentic), forgery, and disguised (when author tries to show his signature as forgery [7]). Researchers use different terms to define forgery types. For instance in [8], forgery is divided into: simple (when forger has no attempt to mimic a signature), random (when forger uses his signature instead of genuine signature), and freehand or skilled (when forger tries to simulate genuine signature as close as possible). In addition, some papers define traced forgery as a sample made by tracing a signature [9, 10]. In [11], simple forgery is defined as a forgery made by ordinary people, while skilled forgery is the result of expert's effort. In this paper, we opt to divide forgery into random and skilled ones, which is the more common categorization in the literature. A sample is random forgery when it is completely dissimilar to the genuine signature -either signed without having the genuine sample or when it is a genuine signature of another author. A sample is skilled forgery when ordinary people put remarkable effort to forge a signature by looking at the genuine sample(s).

In [12], SVS considered as an application of handwriting recognition. In [8] and [13], existing SVS are surveyed up to 1989 and 1994, respectively. More recent developments may be tracked in [10], [6], and [14].

Datasets are vital parts of SVSs. They are prerequisites for training classifiers as well as evaluation and comparison of different SVSs. Therefore, collecting datasets strongly advances the field. The first step is to collect signatures of the people in the community that the verification system is designed for. Signature styles are different in distinct cultures [15]. For instance, while English signatures usually consist of reshaped handwritten names, Persian signatures are often cursive and independent of the names [16]. As a result, a Persian SVS requires a Persian signature dataset.

Signature datasets must be rich. The term "rich" for signature datasets refers to the number of samples and participants, as well as the variables involved in collection procedure, which include signing period, writing instrument, paper, provided space for signing, samples showed to the forgers, the efforts of the forgers, meta-data, etc. These variables come from daily life, for instance, signature of a person changes by the time, by the used pen or by the limited space for signing. To create a more realistic dataset, these variables should be considered as much as possible.

In the offline SVS literature, datasets are mainly collected in Western, Chinese, or Japanese societies. Samples of these datasets differ significantly from the Persian signatures, and consequently cannot be used for Persian SVSs. To our knowledge, there is merely one Persian offline signature dataset that is small and not rich.

In addition to the need for novel and rich culture-based signature datasets, it is strongly essential to define standard experimental setups to provide standard and fair comparison between results of different SVSs. Lack of standard setups in many publicly available datasets accounts for incomparable reported results, mainly due to different training and test conditions (different number of genuine samples in training set, adding or neglecting random forgeries in training set, etc.).

This paper introduces a new and public[1] Persian offline signature dataset, UTSig. This rich dataset consists of significant numbers of classes and samples, where aforementioned variables are considered during signature collection procedure. UTSig provides the research community with the opportunity to train, test, and compare different Persian offline SVSs, and to evaluate different culture-independent classifiers on a rich dataset by using its proposed standard experimental setups.

In this paper, we compare UTSig with the other public datasets in terms of considered variables. We show the distinct characteristics of Persian signatures in terms of number of branch points and end points. We propose and evaluate four standard experimental setups for UTSig. To examine the performance of a common SVS on the public datasets in similar conditions, we utilized same setup on UTSig and other datasets.

The rest of this paper is organized as follows: Section 2 reviews popular offline signature datasets. Section 3 introduces the new Persian offline signature dataset. Section 4 proposes experimental setups. In Section 5, offline signature datasets are compared. Experiments and results are presented in Section 6. The paper is concluded in Section 7.

## 2. State of the Art in Offline Signature Datasets

This section reviews the main characteristics of popular offline signature datasets in the literature.

Spanish dataset MCYT-75 [17], a sub-corpus of MCYT bimodal database [18], has 75 classes containing 15 genuine and 15 forged signatures contributed by 3 user-specific forgers. In this dataset, individuals used inking pen and paper over a pen tablet. Forgers had genuine images and imitated their shapes and natural dynamics.

GPDSsignature [19] has 160 classes with 24 genuine samples gathered in a single day and 30 forged samples imitated by 10 forgers form 10 genuine specimens. To make forgeries, forgers had a random genuine sample and enough time. Individuals used black or blue ink, and white papers with 2 different box sizes.

GPDS-960 [11] contains 960 classes with 24 genuine and 30 forged signatures for each one. Genuine samples were signed in a single day on papers with 2 different box sizes. Totally, 1920 individuals apart from genuine persons made forgeries, and each forger had 5 genuine samples from 5 specimens (one sample per specimen).

ICDAR2009 signature competition [2] offline dataset contains training and evaluation sets. For training, NISDCC signature collection acquired in WANDA project [20] was used. It has 12 classes, and each class contains 5 genuine and 5 forged samples. 31 forgers made forgeries. Evaluation set was collected in the Netherlands Forensic Institute (NFI). It contains 100 classes and each class has 12 genuine and 6 forged samples made by 4 forgers from 33 writers.

---

[1] UTSig Dataset is freely available at MLCM lab website: http://mlcm.ut.ac.ir/Datasets.html

Pourshahabi et.al. [21] employed a Persian signature dataset (FUM). It has 20 classes, and each class contains 20 genuine and 10 forged signatures. Further information about this dataset is unavailable.

In 4NSigComp2010 [7] La Trobe signature collection is used. Its training set consists of 9 reference signatures by one author and 200 questioned signatures including 76 genuine, 104 simulated (forged) signatures made by 27 freehand forgers, and 20 disguised samples. Genuine and disguised samples were signed over a week. In addition, the author wrote another 81 genuine signatures for forgery sample collection. To make forgeries, forgers had 3 out of 81 samples, and imitated without tracing in two ways: forging 3 times without practice and simulating 3 times after 15 practices. The test set has 25 signatures of another person written during 5 days and 100 questioned samples including 3 genuine, 7 disguise, and 90 simulated (forged) signatures written by 34 freehand forgers from lay persons and calligraphers. All individuals used ball-point pen on same papers.

SigComp2011 [3] offline dataset contains Chinese/Dutch signatures, where the training set consist of 235/240 genuine and 340/123 forged signatures from 10/10 different Chinese/Dutch authors. Test sets consist of 116/648 references, 120/648 questioned, and 367/638 forged samples from 10/54 different Chinese/Dutch authors.

In 4NSigComp2012 [22] a new dataset is introduced that contains 3 authentic authors with 15 to 20 references and 100 to 250 questioned signatures. For each class, questioned signatures contain 20 to 50 genuine, 8 to 47 disguised, and 42 to 160 forged samples. Genuine and disguised samples were collected during 10 to 15 days. The number of forgers varied from 2 to 31. Each forger was provided with 3 to 6 authentic samples. Forgers used pen and pencil, and forged with and without practice.

In SigWiComp2013 [5] new Dutch and Japanese offline signature datasets are introduced. Japanese dataset is converted from online signatures that contains 30 classes with 42 genuine samples per class, made in 4 days, and 36 forgeries made by 4 forgers. Dutch dataset has 27 authentic persons who made 10 signatures with arbitrary writing instruments during 5 days. For forgeries, 9 persons used any to all of the supplied specimen signatures as model(s). In average, there are 36 forgeries for each class.

In existing datasets, there is merely one small Persian dataset without any description about data collection procedure. Therefore, study of the Persian offline SVSs demands for a new rich dataset. Among non-Persian datasets, MCYT-75 and GPDS-960 are more commonly used in the literature, while MCYT-75 has relatively small number of genuine and forged samples per class, and GPDS-960 with the largest number of classes, suffers from small period of genuine samples collection. Moreover, GPDS-960 is not publicly available to the research community.

### 3. UTSig Dataset

UTSig (University of Tehran Persian Signature) dataset consists of 8280 images from 115 classes. Each class belongs to one specific authentic male person and has 27 genuine and 45 forged samples of his signatures. Fig.1 shows samples from 4 classes. We randomly selected participants from Iranian undergraduate and graduate students of University of Tehran and Sharif University of Technology. Their ages were between 18 and 31 (average 24.14) and 90% of them were right-hand writers. Genuine participants consisted of 100% males; however, 40% female and 60% male forgers participated. In other words, 115 authentic persons corresponding to 115 classes were male, but their genuine samples were forged by both genders. Both authentic persons and forgers agreed that their signatures to be published publicly for any non-commercial academic purpose. No identity information is attached to the signatures, and participants have given their consent under condition of anonymity.

Participants signed with arbitrary pens on A4-sized white forms in specific boxes, Fig.2. We scanned the forms with 600 dpi resolution, and stored as 8 bit grayscale TIF files. We manually removed considerably large artefacts (e.g. artefacts from bad printing), and used a simple noise removal where pixels brighter than a threshold were assigned to pure white (255 in grayscale). To estimate the threshold, we scanned five blank papers and found the darkest pixel, which resulted in 237 in grayscale as threshold.

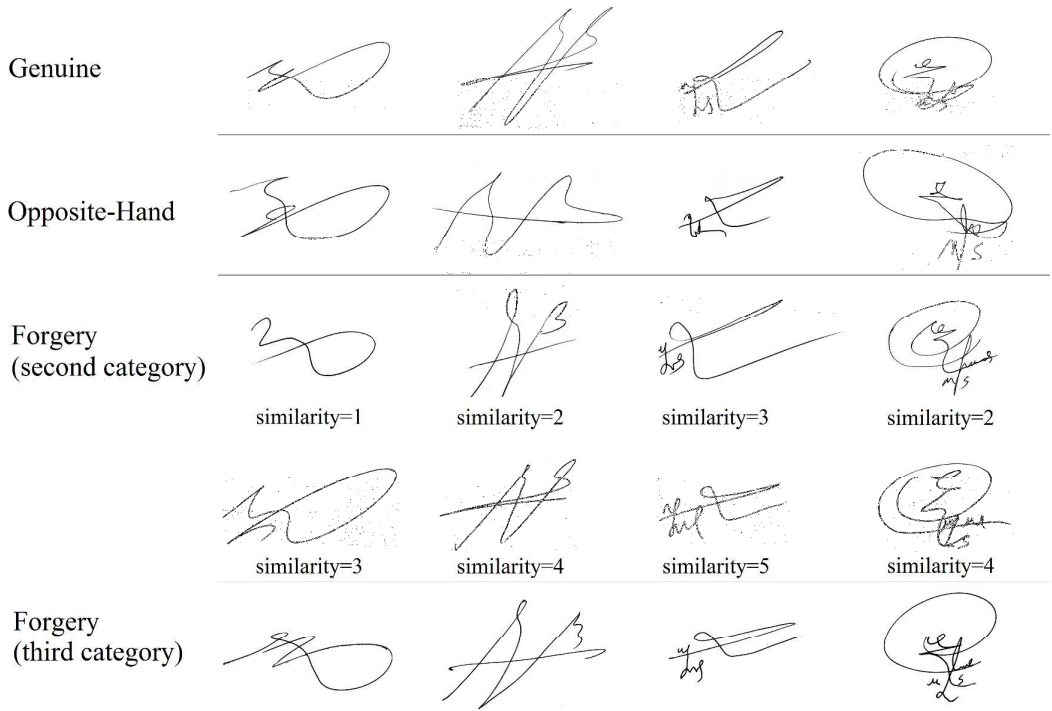

*Fig. 1.* *Four genuine samples from UTSig, their opposite-hand and forgeries (similarity scores were determined by forgers)*

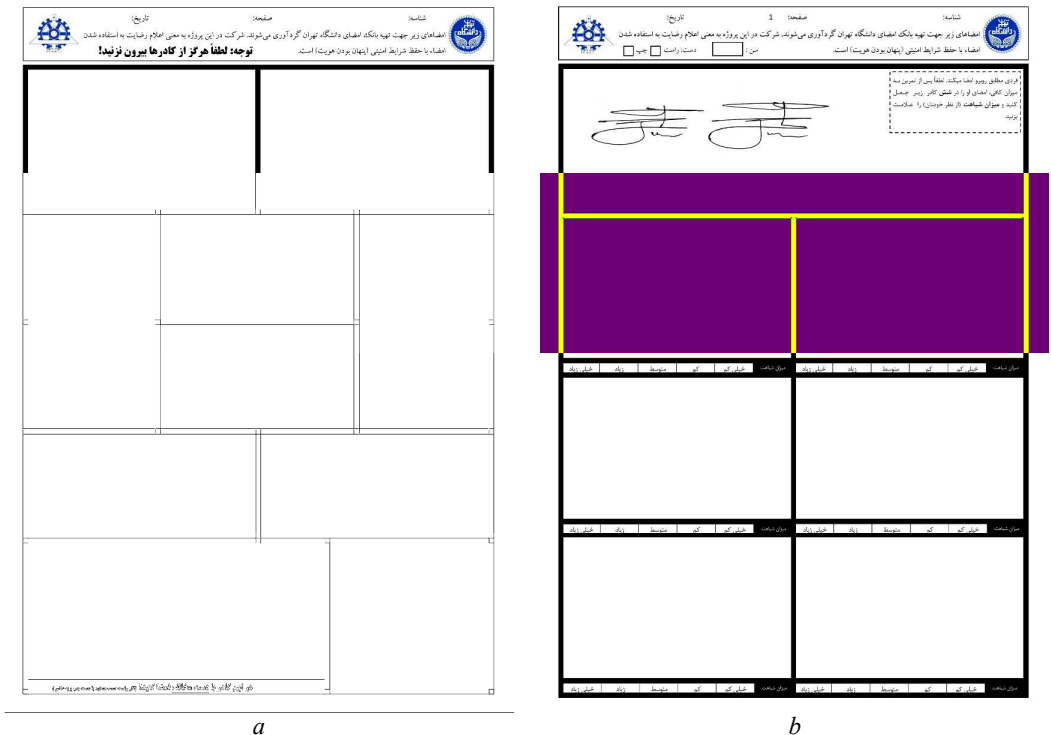

*Fig. 2.* *Data collection forms*
a Genuine and opposite-hand form
b Forgery form

### 3.1. Genuine Signatures

To obtain genuine signatures, 115 male participants signed 10 times on a form and repeated the action for 3 days. In each day, first 9 signatures were genuine and the last one was an opposite-hand signature, signed by the same writer, that can be used as forgery or disguise. As a result, we collected 3105 genuine and 345 opposite-hand signatures. We consider opposite-hand signed samples as forgery, since they lack many genuine traits.

The genuine data collection form, Fig. 2a, contained 10 boxes in 6 different sizes: 9 for genuine and 1 for opposite-hand signatures. The reason for using different sizes is to provide natural conditions or constraints that occur in public service application forms and cause authentic changes that consequently affect the accuracy of SVSs [23]. To keep the nature of signature unchanged, writers were allowed to sign either vertically or horizontally. Furthermore, the box sizes were large enough for Persian signatures. However, whenever a signature crossed the box boundaries, we gave a fresh form to the genuine author.

### 3.2. Forged Signatures

UTSig consists of 5175 forgeries divided into 3 categories. The first category contains 345 opposite-hand signatures of the authentic authors -three sample per class for 115 classes. We define opposite-hand samples as forgery, because they lack many of the genuine traits of genuine samples and cannot be considered as genuine signatures. According to the definition of random forgery and skilled forgery, opposite-hand samples are not random forgery since their overall shapes are not completely dissimilar to the genuine samples, therefore we consider them as skilled forgery. Note that, skilled forgery occurs when a sample is similar to the genuine samples but lacks all the authentic traits. Similarly, opposite-hand signatures are made by the authors who know all the details of their signatures but these samples lack all the traits since physical ability of the opposite-hand is dissimilar to the genuine hand. In real conditions (e.g. signature-based identity verification in banks), collecting skilled forged samples from forgers is impractical, and consequently training SVSs by using such skilled forgeries is impossible. Therefore, we use opposite-hand samples as skilled forgery to enhance the performance. We justify this practice in Section 6.2.3.

The second category contains skilled forged samples obtained from 230 persons apart from authentic ones. They were asked to make signatures as similar as possible to genuine samples. Each forger was provided with three forgery forms from three different classes, where their observable genuine samples on each form varied from one to three signatures of the same class. They were free to practice as much as they want, before filling out the forms. When a forger had more than one genuine sample, it was up to her/him whether to use one or more samples. Each forger made six forgeries on each forms, in boxes with the same size, Fig.2b. We considered six forgery forms per class that were given to six different forgers, where two skilled forgers saw one genuine sample, two skilled forgers saw two genuine samples, and two skilled forgers saw three genuine samples.

For each forged sample, the forger was asked to determine its similarity to genuine sample(s), Fig.1, from very low to very high at five levels (1, 2, 3, 4 or 5). The average of obtained similarity scores is 2.72 (2.67, 2.71 and 2.79, when 1, 2, and 3 genuine sample were observable, respectively). In sum, this category has 4140 skilled forged samples, where 91% of the samples (74% of total forgeries) were ranked by the forgers.

The third category, which has 690 samples, also contains skilled forgery, but this time the samples were forged by a more skilful person, as compared with the two previous categories. The used form was same as the second category and the observable sample for this category was only one random genuine sample.

In all categories, the writers were ask to be cautious about box boundaries. However, we manually refined samples that crossed the boundaries (less than 1% of the forgeries needed this treatment). Fig.3 shows a sample before and after manual refinement. In the refinement procedure, similar to [11], we removed black lines of boxes in such a way that the curve of the signature remains visually natural in and around the crossing point. This procedure does not mean to be perfect, but effective and practical.

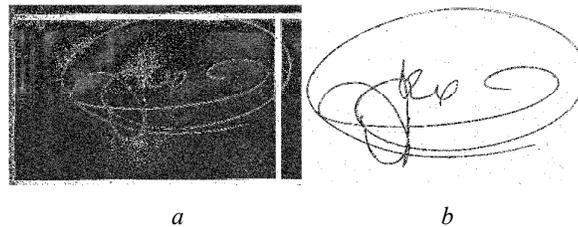

*a*          *b*

***Fig. 3.*** *An example for manual refinement*
a Original sample which crossed the boundaries
b Refined sample

## 4. Proposed Experimental Setups

Writer-Independent (WI) and Writer-Dependent (WD) are the two main approaches in SVSs. In WI, to estimate the distribution of within-writer and between-writer similarities, all genuine and forged samples are compared pair-to-pair and regardless of their authors. In testing phase to verify a questioned signature, it is checked in the both distributions. WD uses author-based samples. In other words, training phase is done separately for each authentic person by samples of his class [24].

### 4.1. Training and Testing Setups

In this paper, we focus on WD and propose four different training and their corresponding testing setups for UTSig dataset. Note that, using skilled forgeries in training set is not realistic, because in real conditions (e.g. signature-based identity verification in banks), it is impractical to collect skilled forgery for new users. Therefore, the proposed setups consist of genuine signatures, random forgeries, and opposite-hand samples.

Genuine vs. random forgery (setup 1): training SVS for each author by using 12 randomly selected of his genuine samples and 5 random forgeries from each other classes ($570 = 5 \times (115 - 1)$). This is a common setup in the literature and used in many papers such as [25] and [26].

Genuine vs. random forgery and opposite-hand (setup 2): training SVS for each author by using 12 randomly selected of his genuine samples, 5 random forgeries from each other classes, and all his opposite-hand samples. This setup can be employed merely for datasets with opposite-hand or disguise samples. As mentioned, it is not recommended to use skilled forgeries in training phase; meanwhile, random forgeries are significantly different from skilled forgeries that SVSs encounter in testing phase. As a result, it is assumed that opposite-hand samples can enhance SVSs.

Genuine vs. opposite-hand (setup 3): training SVS for each author by using 12 randomly selected of his genuine samples and all his opposite-hand samples. This new setup may improve the performance, but it should be used with a classifier that is suitable for small sample size problems, because there are only 12 positive and 3 negative training samples.

Genuine alone (setup 4): training SVS for each author by using 12 randomly selected of his genuine samples. This is a common setup in the literature too and used in many papers such as [19] and [27].

We propose to test the system with 15 remaining genuine samples, along with remaining skilled forgeries and random forgeries. Note that, we consider all the remaining samples of the other classes as random forgeries. For instance, for setup 1, it consists of $7638 = (72 - 5) \times (115 - 1)$ samples. Table 1 shows suggested training and testing setups for UTSig dataset. This table shows proposed setups for each class. Therefore, to train and test all classes it must be repeated for all 115 classes.

**Table 1** Proposed training and testing setups for each class of UTSig.

| Setup   | Training Setup                              | Testing Setup                          |
|---------|---------------------------------------------|----------------------------------------|
| Setup 1 | 12 Genuine + 570 Random                     | 15 Genuine + 45 Skilled + 7638 Random  |
| Setup 2 | 12 Genuine + 570 Random + 3 Opposite-hand   | 15 Genuine + 42 Skilled + 7638 Random  |
| Setup 3 | 12 Genuine + 3 Opposite-hand | 15 Genuine + 42 Skilled + 8208 Random |
| Setup 4 | 12 Genuine | 15 Genuine + 45 Skilled + 8208 Random |

*4.2. Evaluation*

In this paper, we follow the paradigm shift introduced in SigComp2011 [3] to use both decision and likelihood based criteria for SVSs. For decision based criteria, we calculate False Acceptance Rate (FAR) (i.e. the percent of forged samples that are incorrectly accepted), separately for random and skilled forgeries, False Rejection Rate (FRR) (i.e. the percent of genuine samples that are incorrectly rejected), and Equal Error Rate (EER) (i.e. the rate at which skilled forgery FAR and FRR are equal). For likelihood criteria, we refer to two information-theoretic measures: cost of log-likelihood-ratio ($\hat{c}_{llr}$) and its minimal possible value ($\hat{c}_{llr}^{min}$). $\hat{c}_{llr}$ is a positive unbounded measure that is calculated by normalizing the weighted summation of $log = (1 + \exp(y))$ and $log = (1 + \exp(-y))$, where $y$ is the classifier output score. $\hat{c}_{llr}^{min}$, the criteria designed for final evaluation, is the minimum or optimized value of $\hat{c}_{llr}$ that is bounded between 0 and 1. The values of these two criteria are affected by the probability scores produced by the SVS for questioned samples. In other words, the better value is obtained if the SVS assigns higher scores for its true and lower scores for its false rejections and acceptances. Further details and mathematical definitions are provided in the original paper [28].

To compare the performance of different SVSs, we use both EER and genuine versus skilled forgery $\hat{c}_{llr}^{min}$, because EER is a standard criterion in the literature and $\hat{c}_{llr}^{min}$ has been proposed in recent signature competitions (4NSigComp2012 [22] and SigWiComp2013 [5]) to evaluate SVSs. As a result, SVSs with less value of EER or $\hat{c}_{llr}^{min}$ have better performance in terms of EER or $\hat{c}_{llr}^{min}$, respectively. In SigWiComp2013, it was shown that good EER does not necessarily mean good $\hat{c}_{llr}^{min}$ [5]. Final results should be the average of ten independent and new experiments. In each experiment, each criterion is calculated once and not separately for each class.

## 5. Comparison of datasets

Some datasets with few changes are publicly available. Table 2 compares UTSig and other public datasets in terms of statistics and variables considered in data collection procedures. UTSig dataset in comparison with other datasets, including the only existing Persian offline signature dataset (FUM), has larger numbers of classes, total samples, and forgers. Although some datasets surpass UTSig in terms of average number of genuine and forged signatures per class, their very small number of authentic authors reduce their usefulness. Meanwhile, UTSig surpasses the other datasets in terms of numbers of forgers, different box sizes, different observable samples, and meta-data (self-score).

**Table 2** Comparison between public offline signature datasets

| Dataset | Authentic Authors | Ave. Genuine/Author | Ave. Skilled Forgery/Author | Ave. Disguised (Opposite-hand)/Author | Total Samples | Number of Forgers | Period of Collecting Genuine Samples (days) | Number of Different Box Sizes | Number of Different Observable Samples for Forgers | Arbitrary Pen | Opposite-hand or Disguise Samples | Meta-Data (e.g. Self-Score) |
|---|---|---|---|---|---|---|---|---|---|---|---|---|
| MCYT-75 [17] [18] | 75 | 15 | 15 | 0 | 2250 | 75 | NA | 1 | NA | NA | No | No |
| ICDAR2009 [2] | 91 | 11 | 27 | 0 | 3462 | 64 | NA | NA | NA | NA | No | No |
| FUM [21] | 20 | 20 | 10 | 0 | 600 | NA | NA | NA | NA | NA | No | No |
| 4NSigComp2010 [7] | 2 | 57 | 97 | 14 | 335 | 61 | 5-7 | NA | 1 | No | Yes | No |
| SigComp2011 Dutch [3] | 64 | 24 | 12 | 0 | 2295 | NA | NA | 1 | NA | NA | No | No |
| SigComp2011 Chinese [3] | 20 | 24 | 34 | 0 | 1176 | NA | NA | 1 | NA | NA | No | No |
| 4NSigComp2012 [22] | 3 | 55 | 91 | 21 | 501 | 39 | 10-15 | 1 | 2 | No | Yes | No |
| SigWiComp2013 Dutch [5] | 27 | 10 | 36 | 0 | 1241 | 9 | 5 | NA | NA | Yes | No | No |
| SigWiComp2013 Japanese [5] | 20 | 42 | 36 | 0 | 1566 | 4 | 4 | NA | NA | NA | No | No |
| UTSig | 115 | 27 | 42 | 3 | 8280 | 230 | 3 | 6 | 3 | Yes | Yes | Yes |

## 6. Experiments

### 6.1. Persian Signatures Characteristic

Signatures in distinct cultures have different shapes. As a result, it seems clear to use different feature extraction and classification methods for the signatures of distinct cultures. For instance in [29] and [30], two-stage approach are used to improve verification accuracy of multi-scripts signatures. First, an identification system finds whether questioned signature is Hindi or English, then different SVSs are used for each type of signatures.

To show signatures differences in available datasets statistically, we used morphological operations to count branch points and end points of each sample. A branch point is a point that the signature crosses itself, and an end point is the beginning and final point of each connected component. First, we binarized genuine samples with a threshold that was approximately the darkest pixel of blank areas of images. Then, we used connected-component analysis, to remove components with fewer than ten pixels from binary images. To refine samples, after applying horizontal dilation operator, we set a pixel to black if five or more pixels in its 3-by-3 neighbours were black (i.e. majority operation). We applied standard morphological skeletonization [31], and finally extracted end points and branch points of skeletonized images by finding pixels with merely one neighbouring pixel (i.e. morphological end-point operator) and finding pixels with more than two neighbours (i.e. morphological branch-point operator), respectively. In order to remove unreal points caused by skeletonization operation, if the Euclidean distance between two or more adjacent branch points or end points was less than a threshold (i.e. 10), we only preserved one point. Fig. 4 shows two branch points and two end points.

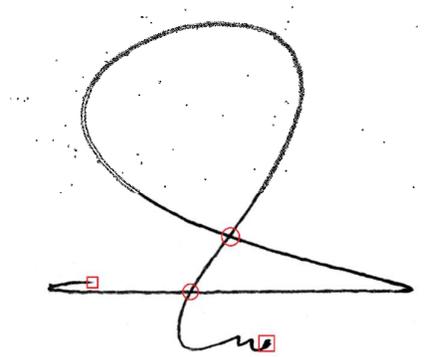

*Fig. 4.* *A signature with two branch points (circles) and two end points (squares)*

Analysis on the number of branch points and end points shows that Persian signature in UTSig and FUM [21] datasets have fewer branch points and end points in comparison with the other datasets, including Dutch, Spanish, Chinese and Japanese ones (see Fig. 5).

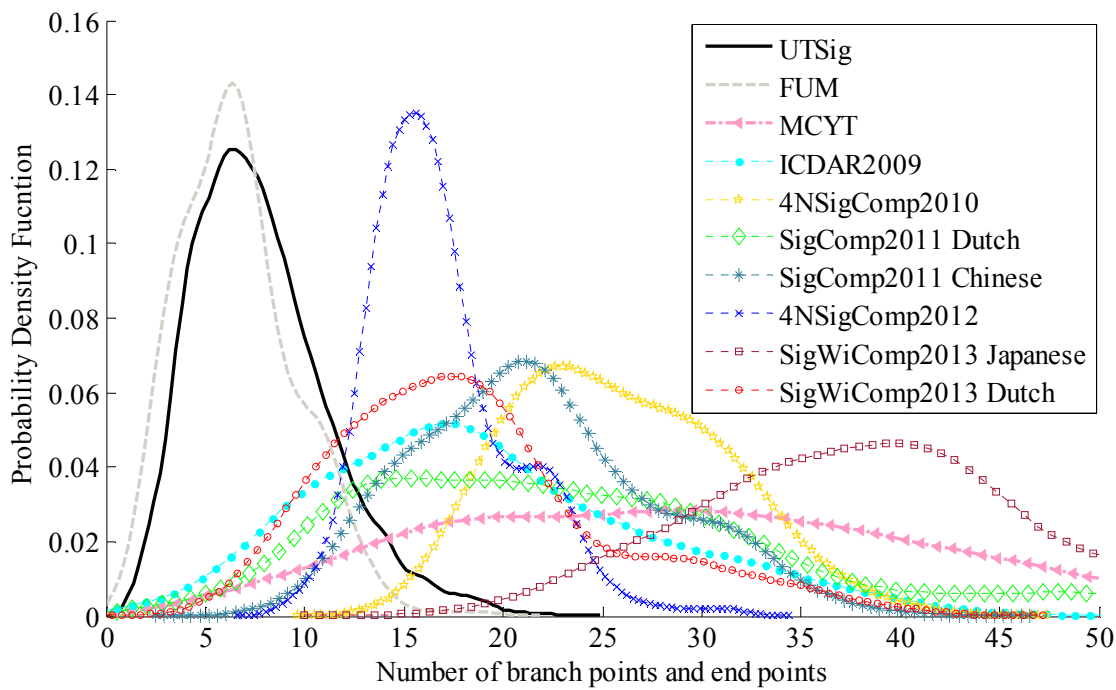

*Fig. 5.* *Estimated probability density function (PDF) of number of branch points and end points in offline signature datasets*

### 6.2. Verification System

*6.2.1 Classifier:* Support Vector Machine (SVM) is a statistical approach to design a supervised classifier. It was originally developed for two-group classification problems. To separate two classes, SVM uses kernels mapping input vectors to a high-dimension feature space, and then constructs a linear decision surface in the high-dimension space [32]. We used SVM to separate genuine and non-genuine samples of each authentic individual. We used linear kernel, which multiplies feature dimensions to build high-dimension space ($K(x_1, x_2) = x_1^T x_2$), for the first, second, and third setups. For the fourth setup, which is a one-class problem, we used one-class SVM with radial basis functions (RBF) kernel, which is defined as $K(x_1, x_2) = \exp(-\frac{||x_1-x_2||^2}{2\sigma^2})$. SVM produces labels that determine the class of the input vector, and in order to calculate EER and construct likelihood ratio, we mapped SVM outputs into probabilities or scores by the method proposed in [33].

*6.2.2 Feature Extraction:* We used fixed-point arithmetic, which is described in [19] as "description of the signature envelope and the interior stroke distribution in polar and Cartesian coordinates". In fixed-point arithmetic feature extraction by using the geometric centres of samples, three parameters are calculated in polar coordinate: derivative of radius of signature envelope, its angle, and the number of black pixels that the radiuses cross when rotate from one point to the next point. In Cartesian coordinates, height, width and the number of transitions form black to white or white to black pixels of signatures are calculated with respect to their geometric centres [19].

*6.2.3 Results:* To find numerical results of UTSig dataset, we repeated experiments for each author 10 times and averaged their results. Results are available in Table 3. In terms of both EER and $\hat{c}_{llr}^{min}$, setup 2 has the best performance, after that in descending order setups 1, 4 and 3 have better results. To find the best result statistically, we used t-test that indicates with a 90% confidence interval, setup 2 surpasses setup 1 and the other setups in terms of EER and $\hat{c}_{llr}^{min}$. Therefore, results with 90% confidence interval justify that using opposite-hand signatures along with random forgeries improves the performance. Hence, collecting opposite-hand samples in future datasets and real conditions can be promising.

Note that, in the fourth setup, according to the nature of one-class SVM, a threshold must be selected. In this paper, we selected one threshold for all authors but better results may be obtained by using user-based thresholds.

**Table 3** Setups results for UTSig including 90% confidence interval.

| Setup | EER | Genuine FRR | Skilled FAR | Random FAR | $\hat{c}_{llr}$ | $\hat{c}_{llr}^{min}$ |
|---|---|---|---|---|---|---|
| 1 | 29.71% ±0.29 | 39.27% | 21.29% | 0.08% | 0.996 | 0.819±0.004 |
| 2 | 29.33%±0.22 | 41.70% | 18.34% | 0.07% | 0.995 | 0.813±0.002 |
| 3 | 34.14%±0.30 | 0.02% | 93.23% | 89.53 % | 1.43 | 0.896±0.002 |
| 4 | 32.46%±0.34 | 32.50% | 32.43% | 4.33% | 2.65 | 0.868±0.003 |

According to the participant self-scores in forgery process, the behaviour of the trained systems has a trend toward distinguishing the low score samples more than the high score samples. The average score of true negative samples is 2.65 while the number for false positive is 2.86.

To check the results with respect to the number of samples or random forgeries in training phase, we repeated the first setup. Table 6 shows that if the number of genuine samples in training set increases, EER and FRR decrease and skilled forgery FAR increases. Meanwhile, the addition of more random forgeries in training set increases EER and FRR, but decreases skilled forgery FAR.

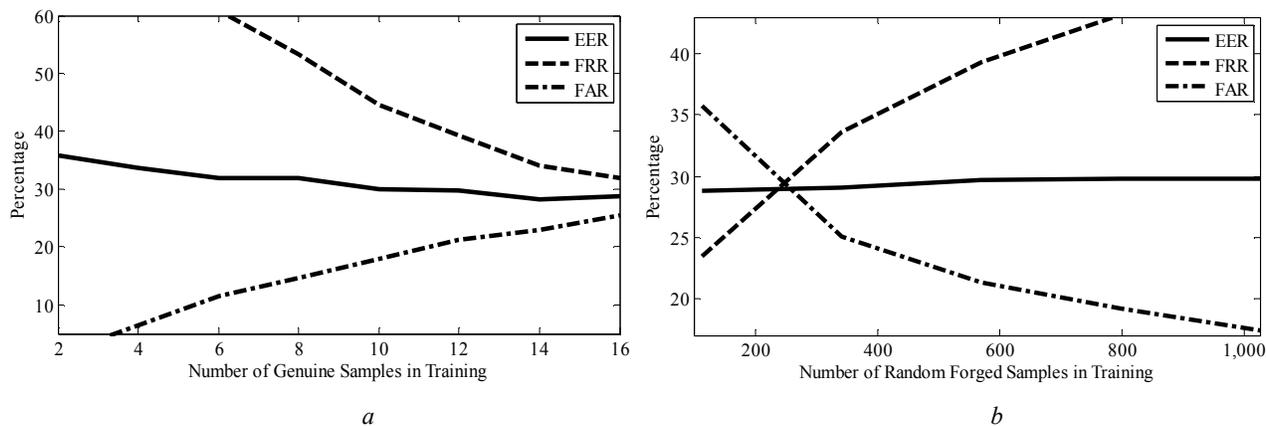

**Fig. 6.** *FRR and FAR variation by*
a Number of genuine samples in training
b Number of forged samples in training

To check the performance of the similar system on the other datasets, we only performed the fourth setup, since the other setups need a variety of random forgeries (from distinct authors) or opposite-hand samples, which are unavailable in all the datasets. Note that, we ignored some datasets due to unavailability, missed, or few genuine samples. Results of the same verification system on the other datasets are reported in Table 4. According to EER and $\hat{c}_{llr}^{min}$, the performance of the system significantly decreases when it is tested on SigWiComp2013, UTSig, SigComp2011, and 4NSigComp2010.

**Table 4** Datasets results for setup 4 including 90% confidence interval.

| Setup | EER | Genuine FRR | Skilled FAR | Random FAR | $\hat{c}_{llr}$ | $\hat{c}_{llr}^{min}$ |
|---|---|---|---|---|---|---|
| MCYT-75 | 25.9±0.28% | 27.2% | 24.5% | 4.9% | 0.815 | 0.722±0.002 |
| FUM | 26.2±0.27% | 26.2% | 26.1% | 0.22% | 0.810 | 0.741±0.002 |
| 4NSigComp2010 | 29.1±0.30% | 28.2% | 30.1% | 0.10% | 0.771 | 0.782±0.004 |
| SigComp2011 Dutch | 32.0±0.31% | 31.6% | 31.1% | 5.9% | 1.008 | 0.851±0.003 |
| 4NSigComp2012 | 22.3±0.24% | 21.4% | 23.5% | 3.8% | 0.825 | 0.604±0.003 |
| SigWiComp2013 Japanese | 33.1±0.30% | 34.9% | 33.2% | 7.5% | 1.372 | 0.879±0.004 |
| UTSig | 32.46%±0.34 | 32.50% | 32.43% | 4.33% | 2.65 | 0.868±0.003 |

## 7. Conclusion

In this paper, we introduced a new and rich Persian offline signature dataset with 115 classes. Each class consists of 27 genuine, 3 opposite-hand, and 42 skilled forged samples. UTSig surpasses existing Persian signature dataset in sample size and variables considered in data collection procedure. In comparison with other public datasets, UTSig has more samples, more classes, more forgers, more box sizes, and more different observable samples. UTSig has opposite-hand samples and self-score. In the sample collection procedure, individuals used arbitrary pens and made genuine samples in 3-day period. These features make UTSig a useful dataset for Persian SVSs. UTSig can be used in evaluation of culture-independent SVSs, as well.

We proposed four different standard WD training and testing setups for UTSig dataset. We observed that using opposite-hand signatures in the training set enhanced the performance of the SVS. We evaluated setups with SVM classifier and fixed-point arithmetic, where the best obtained performance was EER=29.33% and $\hat{c}_{llr}^{min}$=0.813.

We also counted branch points and end points of signatures in available datasets in different languages, and concluded that Persian signatures have relatively fewer numbers of branch points and end points.

## 8. Acknowledgments